\documentclass{article}
\usepackage{spconf,amsmath,graphicx}                                                      
\usepackage{float} 
\usepackage{subfig}
\usepackage[export]{adjustbox} 
\usepackage{float} 
\usepackage{bbding}
\usepackage{booktabs}
\usepackage{multirow}
\usepackage{amssymb}
\usepackage{threeparttable}
\usepackage{stfloats}

\title{Self-supervised Facial Action Unit Detection with Region and Relation Learning}
\name{Juan Song, Zhilei Liu\sthanks{Corresponding author, Email: zhileiliu@tju.edu.cn}}
\address{College of Intelligence and Computing, Tianjin University, Tianjin, China}

\begin{document}
\ninept
\maketitle
\begin{abstract}
Facial action unit (AU) detection is a challenging task due to the scarcity of manual annotations. Recent works on AU detection with self-supervised learning have emerged to address this problem, aiming to learn meaningful AU representations from numerous unlabeled data. However, most existing AU detection works with self-supervised learning utilize global facial features only, while AU-related properties such as locality and relevance are not fully explored. In this paper, we propose a novel self-supervised framework for AU detection with the region and relation learning. In particular, AU related attention map is utilized to guide the model to focus more on AU-specific regions to enhance the integrity of AU local features. Meanwhile, an improved Optimal Transport (OT) algorithm is introduced to exploit the correlation characteristics among AUs. In addition, Swin Transformer is exploited to model the long-distance dependencies within each AU region during feature learning. The evaluation results on BP4D and DISFA demonstrate that our proposed method is comparable or even superior to the state-of-the-art self-supervised learning methods and supervised AU detection methods.
\end{abstract}
\section{Introduction}
\label{sec:intro}
Facial Action Coding System (FACS)~\cite{ekman1978facial} defines a unique set of non-overlapping facial muscle actions known as Action Units (AUs), the combination of the occurrence or absence of each AU can achieve almost any facial expression. Due to the objectivity of action units, AU detection has drawn significant interest and has been widely applied in human-computer interaction, affect analysis, mental health assessment, etc. However, current research on AU detection is mostly implemented based on supervised learning methods, which rely on large-scale AU-labeled images.

Supervised learning-based AU detection methods tend to mine AUs' characteristics to obtain more discriminative AU features, such as localization and correlation. Several works are based on those attributes to improve the accuracy and robustness of the model. For example, Zhao et al.~\cite{zhao2016deep} proposed a locally connected convolutional layer that learns region-specific convolutional filters from sub-areas of the face. JAA-Net~\cite{shao2018deep, shao2021jaa} integrated the facial alignment task with AU detection to train the network to learn better local features. To exploit the correlation among AUs, Niu et al.~\cite{niu2019local} proposed the LP-Net to model person-specific shape information to standardize local relationship learning. Li et al.~\cite{li2019semantic} proposed SRERL to embed the relation knowledge in regional features with a gated graph neural network. Recently, the transformer has also improved the performance of the model~\cite{yan2021self,jacob2021facial}. These works benefit from leveraging more AU-related qualities.
However, supervised work depends on a considerable number of accurately labeled images. Since labeling AUs is time-consuming and error-prone, the lack of available AU-labeled data limits their generalization. In contrast, unlabeled face data is easy to collect and available in large quantities.

In this paper, discriminative AU representations are learned using the self-supervised method to decrease the demand for AU annotations. To extract its supervised knowledge from vast amounts of unlabeled input, self-supervised learning (SSL) mostly requires auxiliary tasks. It demonstrates excellent promise in applications like object detection and image classification
~\cite{caron2021emerging,chen2021exploring,roh2021spatially}.
Several recent works have attempted to use self-supervised learning to boost the accuracy of AU detection~\cite{wiles2018self,chang2022knowledge,li2019self,lu2020self,yan2021self}. Wiles et al.~\cite{wiles2018self} proposed FAb-Net to utilize the facial movement's transformation between two adjacent frames as the supervisory signal and learn the facial embedding by mapping the source frame to the target frame through reconstruction loss. Inspired by FAb-Net, Li et al.~\cite{li2019self} respectively change the facial actions and head poses of the source face to those of the target face, decoupled facial features from the head posture. Lu et al.~\cite{lu2020self} leveraged the temporal consistency to learn feature representation through contrastive learning. In other work, self-supervised learning is used as an auxiliary task~\cite{yan2021self}.
However, most of these studies only learn about global facial features and ignore the task-related domain knowledge, such as the unique properties of AU: locality and relevance. Only global features have a limited impact on AU recognition since each AU is associated with one or a small group of facial muscles and does not manifest independently. Besides, several self-supervised methods~\cite{caron2021emerging,chen2021exploring} learn powerful visual representations for single object detection via contrastive learning. However, because the majority of contrastive approaches now in use create self-supervised tasks utilizing the random crop or temporal consistency, they are prone to provide insufficient AU representations and underutilize the advantages of static datasets.

This paper proposes a novel self-supervised framework for facial action unit detection using region and relation learning (RRL). Our proposed framework learns AU representations through a two-step training process. The first phase is self-supervised pre-training, where we investigate three levels of representation while using the special qualities of the AU to understand the features associated with it. Specifically, for each image, we create two different augmented views to guarantee global similarity and employ the Swin Transformer~\cite{liu2021swin} as our backbone to extract long-range dependencies and global features. Then, we propose an AU-related local feature learning method guiding the network to learn a variety of AU features in the facial image. Moreover, an improved optimal transport algorithm is proposed to exploit the relations among AUs, providing a new way for learning the correlation characteristics of AUs. In the transfer learning stage, we train simple linear classifiers upon the self-supervised representations for downstream AU detection tasks. 
\vspace{-2mm}
\section{METHOD}
\label{sec:format}
\begin{figure*}[htbp]
\centering
\vspace{-2mm}
\includegraphics[scale=0.5]{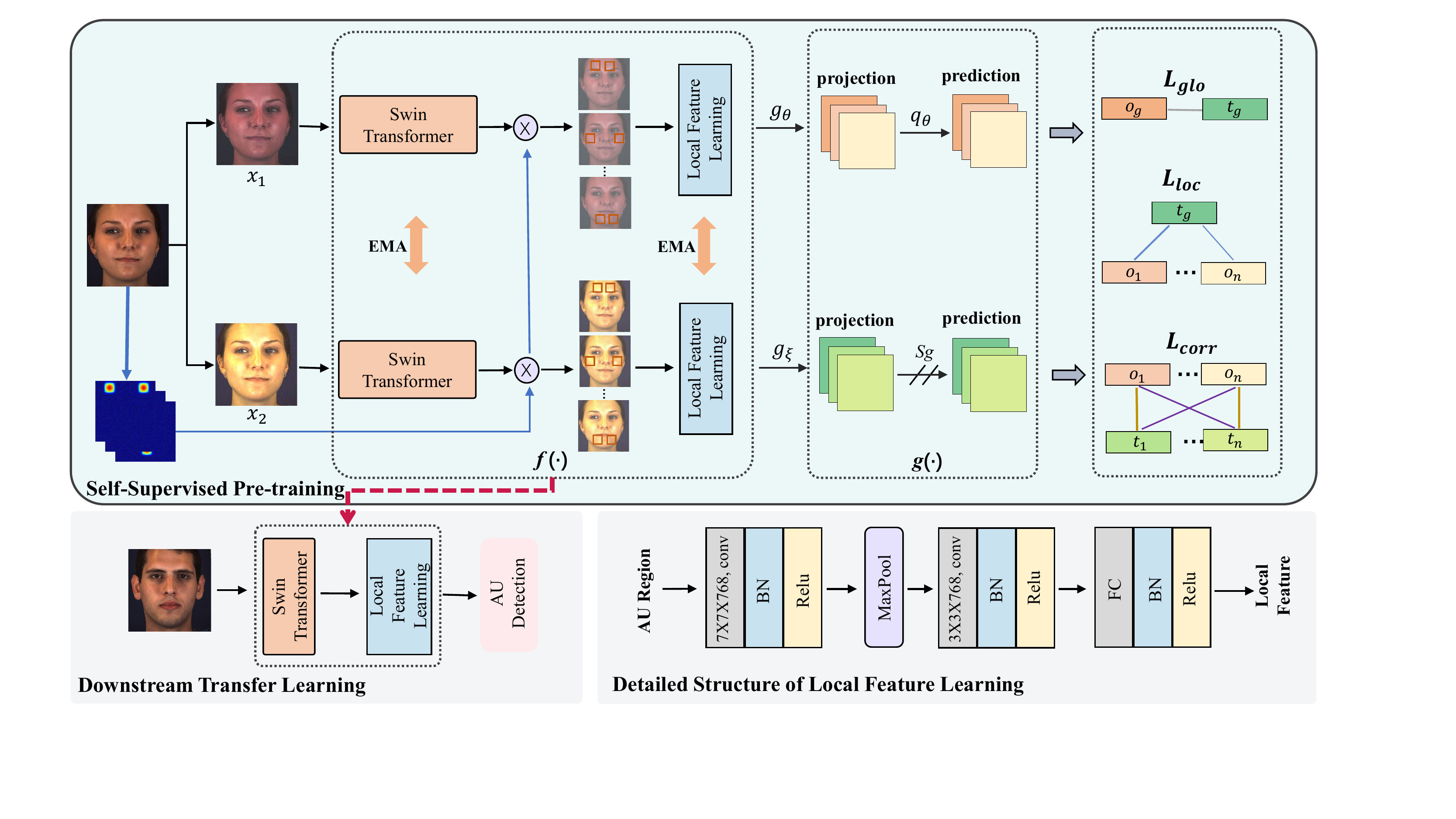}
\caption{The framework of RRL. In the self-supervised pre-training stage, $f(\cdot)$ based on Swin Transformer and a CNN are utilized to extract global and local representations for each augmented view. $g(\cdot)$ maps local facial features to low-dimensional latent space and three components are introduced to train the framework in a self-supervised manner. At the downstream transfer stage, everything but $f(\cdot)$ is discarded. }
\label{fig1}
\vspace{-2mm}
\end{figure*}

\vspace{-2mm}
In this section, we describe the proposed framework in detail. Fig.\ref{fig1} demonstrates the overview of the proposed framework. RRL contains two stages: self-supervised pre-training and downstream transfer learning. The pre-training stage involves three modules: global feature learning, local feature learning, and optimal transport for relation learning. Finally, after pre-training, the learned facial representations will be used for AU detection.

\vspace{-2mm}
\subsection{Global Feature Learning}
Motivated by BYOL~\cite{grill2020bootstrap}, we use two neural networks: the online network which is defined by a set of parameters ${\theta}$ and the target network parameterized by ${\xi}$. The target network provides a regression target to train the online network. The parameters ${\xi}$ of the target network are updated by the online network's parameter ${\theta}$ according to Eq.~\ref{func:1}, in which ${m}$ is the exponential moving average (EMA) parameter with decay rate.

\begin{equation}
\xi=m * \xi+(1-m) * \theta
\label{func:1}
\end{equation}
To model long-distance dependencies, we use Swin Transformer as our backbone. Swin Transformer was selected due to its competitive results in both supervised and self-supervised works, with hierarchical and shifted windowing mechanisms that can help extract more precise facial features and save computational effort. 

Given an input image ${x}$, we first generates two augmented images $x_{1}$ and $x_{2}$.
These augmented images are fed into $f(\cdot)$, which are respectively named by $f_{\theta}$ and $f_{\xi}$. Within the online network, we then perform the projection ${z}_{1}^{g} = g_{\theta}(o_{g})$ from the pooled representation $o_{g}$, followed by the prediction $q_{\theta}({z}_{1}^{g})$. At the same time, the target network outputs the target projection from $t_{g}$ such that ${z}_{2}^{g} = g_{\theta}(t_{g})$. Our global loss is then defined as the cosine similarity between the prediction and the target projection as shown in Eq.~\ref{con:glo}.

\begin{equation}
\mathcal{L}_{\text {glo}}\triangleq-\frac{\left\langle\mathbf{z}_{2}^{g}, q_{\theta}(\mathbf{z}_{1}^{g})\right\rangle}{\|\mathbf{z}_{2}^{g}\|_{2} \cdot\left\|q_{\theta}(\mathbf{z}_{1}^{g})\right\|_{2}}
\label{con:glo}
\end{equation}

\subsection{Local Feature Learning}
\label{sec:typestyle}
As noted in self-supervised learning works, comparing random crops of an image plays a central role in capturing information in terms of relations between parts of a scene or an object~\cite{caron2021emerging}. However, recent work on self-supervised AU detection has not taken this into account. Moreover, the multi-crop strategy in conventional self-supervised object detection could undermine facial AU information. To obtain rich facial information, the primary issue is how to learn AU local features without destroying AU information. Unlike previous methods that random crop patches, RRL uses the attention map to make the model focus more on AU-specific regions, and the local feature learning module is used for AU feature refinement.
                                                
Specifically, We first detect 68 facial landmarks to initialize the attention maps of the AUs for a given input image. Landmarks specific to each action unit are defined similarly to EAC-Net~\cite{li2017eac}. We fit ellipses to landmarks as the initial regions of interest for each AU and smooth the image (Gaussian with $\sigma$ = 3), thus obtaining $K$ AU attention maps $i_{k}$ of 14x14 size, where ${k}$ = \{1,\dots, ${K}$\}. Then, We fuse the features learned by Swin Transformer with the attention maps and input them into the local feature learning network together with the global feature. We use [$o_{1}$,$o_{2}$,\dots,$o_{k}$] for local vectors and $o_{g}$ for global vectors, where the letter $o$ represents the output of the online network and $t$ represents the output of the target network.

Finally, we minimize the cosine distance of the global and local features corresponding to the two networks as shown in Eq.~\ref{con:gloc}.
\vspace{-1mm}
\begin{equation}
\scriptsize
\resizebox{.92\hsize}{!}{
$\mathcal{L}_{\text {loc}}\triangleq-\frac{1}{2K}\sum_{k=1}^{K}\left[\frac{\left\langle q_{\theta}({z}_{1}^{k}), g_{\xi}(\mathbf{t_{g}})\right\rangle}{\| q_{\theta}({z}_{1}^{k})\|_{2} \cdot\left\|g_{\xi}(\mathbf{t_{g}})\right\|_{2}} + \frac{\left\langle g_{\xi}(\mathbf{t_{k}}), q_{\theta}(\mathbf{z}_{1}^{g})\right\rangle}{\|g_{\xi}(\mathbf{t_{k}})\|_{2} \cdot\left\|q_{\theta}(\mathbf{z}_{1}^{g})\right\|_{2}}\right]$
}
\label{con:gloc}
\end{equation}

\subsection{Optimal Transport for Relation Learning.}
\label{sec:majhead}

Similar to~\cite{li2022univip} using Optimal Transport (OT) to learn the discrimination between instances, an improved OT method is proposed here to learn the correlations of AUs.

\begin{figure}[H]
\centering
\includegraphics[scale=0.35]{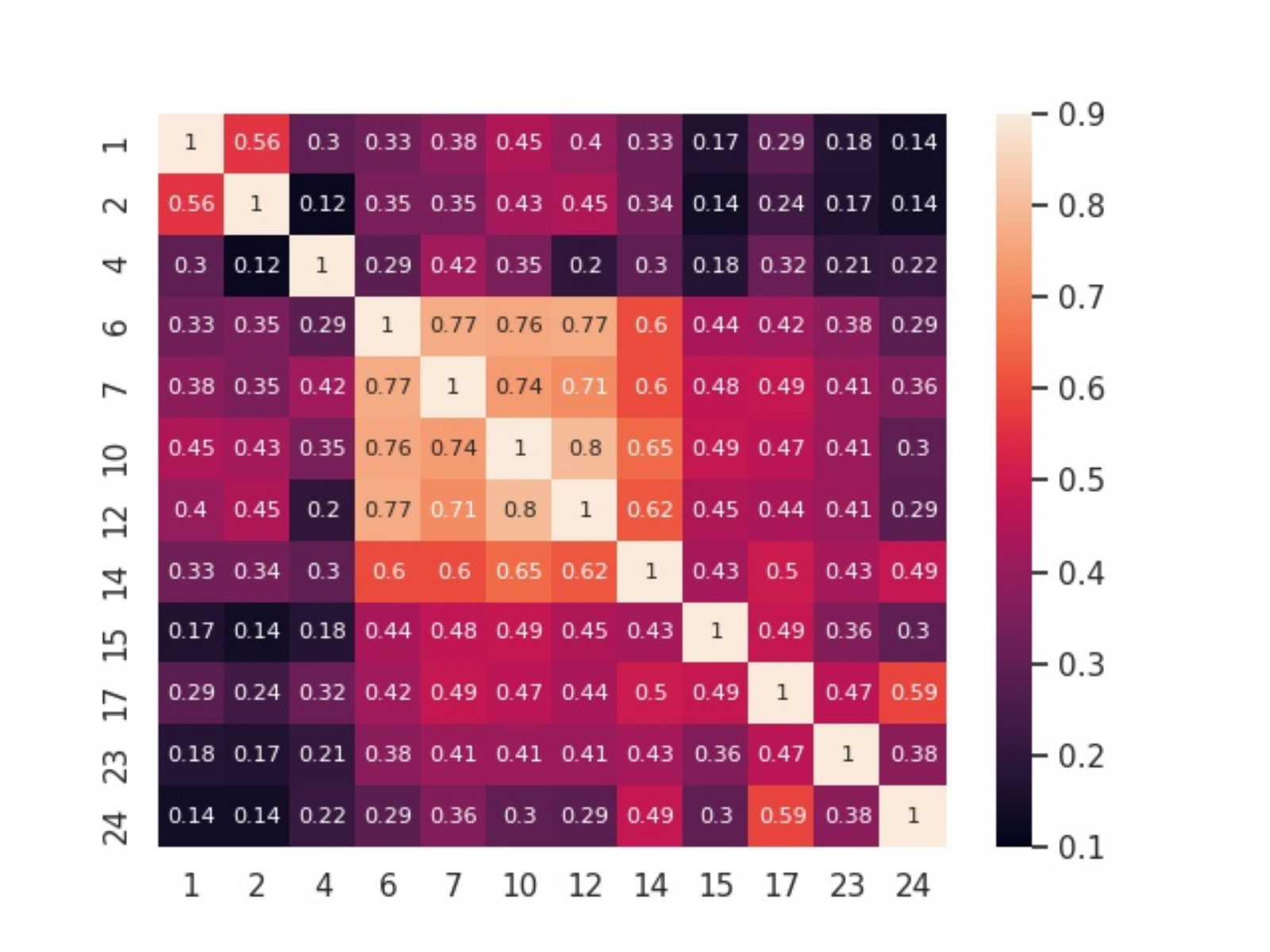}
\caption{Relationship matrix $M_{i j}$ of AUs.}
\label{fig:fig4}
\end{figure}

The AU outputs from the online network [$o_{1}$,$o_{2}$,\dots,$o_{k}$] are considered to be the $k$ goods of the supplier, while vector [$t_{1}$,$t_{2}$,\dots,$t_{k}$] from the target network are considered to be the demands of the demander. In addition, we define the unit transportation cost between the demander and the supplier as the value of the AU relationship matrix. Taking~\cite{liu2020relation} as the reference, we count the correlation between AU labels in the training set and get the AU relationship matrix $M_{i j}$ of AU, as shown in Fig.\ref{fig:fig4}. The larger the matrix value, the greater the correlation between the two AUs. Therefore, the unit transportation cost from the supplier node to the demander node is defined as Eq.~\ref{con:c}.

\begin{equation}
\begin{array}{ll}
{C}_{i j}= 1 - {M}_{i j}
\label{con:c}
\end{array}
\end{equation}  
The correlation degree of each AU pair can be expressed as the optimal matching cost between two sets of vectors. The more relevant the vector, the lower the transport cost.

Following~\cite{li2022univip}, the marginal weights $a_{i}$ and $b_{j}$ are expressed by the product of global vectors and local vectors. In that case, the importance of the AU region depends on its contribution to the whole face as defined in Eq.~\ref{con:b}, where the function max(·) ensures the weights are always non-negative.
\begin{equation}
\begin{aligned}
a_{i} = \left\{\mathbf{o}_{i}^{T} \cdot \mathbf{t}_{g}, 0\right\},\quad b_{j} = \left\{\mathbf{t}_{i}^{T} \cdot \mathbf{o}_{g}, 0\right\}
\label{con:b}
\end{aligned}
\end{equation}
We use the cost matrix $C$ to calculate the optimal transmission scheme ${\pi}^{*}$, and solve the OT problem through the Sinkhorn-Knopp iteration. Then we can define the correlation loss function to learn the distinguishing characteristics as shown in Eq.~\ref{eq:6}.
The loss would be minimized only if the representations of each AU are similar to the associated AU.
\begin{equation}
\label{eq:6}
\mathcal{L}_{\text{corr}}(\mathbf{O}, \mathbf{T}) \triangleq-\sum_{i=1}^{K} \sum_{j=1}^{K} \frac{\mathbf{o}_{i}^{T} \mathbf{t}_{j}}{\left\|\mathbf{o}_{i}\right\|\left\|\mathbf{t}_{j}\right\|} {\pi}^{*}
\end{equation}

After introducing each loss term, the overall loss is defined as following:

\begin{equation}
\mathcal{L}_{\text{all}}={\alpha_{1}}{L}_{\text{glo}} + {\alpha_{2}}{L}_{\text{loc}} + {\alpha_{3}}{L}_{\text{corr}}
\label{con:all}
\end{equation}
where $\alpha_{1}$, $\alpha_{2}$, $\alpha_{3}$ are trade-off parameters. 

\section{Experiments}
\label{sec:print}
\subsection{Experimental Setup}

\textbf{Training.}
We conducted the self-supervised pre-training on the EmotioNet~\cite{fabian2016emotionet} dataset. The \textbf{EmotioNet} is collected in the wild, including a total of 950,000 images. All images are resized to 224 × 224 as the input of the network. Facial landmarks are detected with Dlib. In addition to random horizontal flipping, random grey scaling, and random Gaussian blur, the data augmentation also has random color distortions. Where, the random color distortion with different brightness, saturation, contrast, and hue, augmentation of different values is applied to two views. We employ an AdamW optimizer and the learning rate is $lr = 0.05 * batch size/256$, using a cosine decay learning rate scheduler. The weight decay also follows a cosine decay from 0.04 to 0.4. The trade-off parameters $\alpha_{1}$, $\alpha_{2}$, and $\alpha_{3}$ are empirically set to 0.4, 0.6, and 1.0 respectively. The exponential moving average parameter $m$ is initialized as 0.98 and is increased to 1.0 during training.

\noindent\textbf{Evaluation.}
We then evaluated the methods on BP4D~\cite{zhang2013high} and DISFA~\cite{mavadati2013disfa} datasets. \textbf{BP4D} contains 41 subjects (23 females and 18 males). There are 328 videos with about 146000 frames with available AU labels. \textbf{DISFA} consists of 26 participants. The AUs are labeled with intensities from 0 to 5. The frames with intensities greater than 1 were considered positive, while others were treated as negative. We choose $K$=12 on BP4D, and 8 on DISFA. In the downstream transfer stage, the linear classifier consists of two layers: a batch-norm layer followed by a linear fully connected layer with no bias. The linear classifier was trained with a cross-entropy loss for each AU and adopted the F1 score to evaluate the performance of the method. We also computed the average overall AUs to measure the overall performance.

\subsection{Experimental Results}
We compared RRL with several self-supervised methods and supervised AU detection methods. Table~\ref{DISFA} and Table~\ref{BP4D} report the F1 score on DISFA and BP4D.

\begin{table*}[h!t]
\renewcommand\arraystretch{1.0}
\centering
\scriptsize
\caption{F1 score for self-supervised and supervised methods on DISFA~\cite{mavadati2013disfa} of multiple facial action units (AUs).}
\label{DISFA}
\begin{threeparttable}
\begin{tabular}{@{}c|cc|cccccc@{}}
\toprule
\multirow{2}{*}{AU} & \multicolumn{2}{c|}{Supervised} & \multicolumn{4}{c}{Self-Supervised} \\ \cmidrule(l){2-3} \cmidrule(l){4-8} 
    & DRML~\cite{zhao2016deep}$\ast$     & JAA-Net~\cite{shao2021jaa}$\ast$   & FAb-Net~\cite{wiles2018self} & TCAE~\cite{li2019self} & TC-Net~\cite{lu2020self}$\ast_{k=1}$ & TC-Net~\cite{lu2020self}$\ast$ & RRL(ours) \\ \midrule
 1  & 17.3 & 43.7 & \underline{27.5} & 24.8 & 10.8 & 18.7 & 15.4 \\ 
 2  & 17.7 & 46.2 & 19.6 & 25.5 & 20.7 & \underline{27.4} & 15.9 \\ 
 4  & 37.4 & 56.0 & 28.7 & 37.3 & 43.3 & 35.1 & \underline{49.5} \\ 
 6  & 29.0 & 41.4 & 45.2 & 34.7 & 37.6 & 33.6 & \underline{48.8} \\ 
 9  & 10.7 & 44.7 & 20.9 & \underline{31.1} & 12.2 & 20.7 & 22.1 \\ 
 12 & 37.7 & 69.6 & 65.6 & 59.6 & 68.7 & 67.5 & \underline{70.3} \\ 
 25 & 38.5 & 88.3 & 67.9 & 58.1 & 62.9 & 68.0 & \underline{81.4} \\ 
 26 & 20.1 & 58.4 & 24.0 & 25.2 & 46.2 & 43.8 & \underline{46.8} \\ 
\midrule
Avg. & 26.7 & 56.0 & 37.4 & 37.0 & 37.8 & 39.4 & \underline{43.8} \\  \bottomrule
\end{tabular}
\begin{tablenotes}
\item $\ast$ means that the values are reported in the original papers.
\end{tablenotes} 
\end{threeparttable}
\end{table*}
\textbf{Comparison with self-supervised methods}
We compare our method with self-supervised methods: FAb-Net~\cite{wiles2018self}, TCAE~\cite{li2019self}, and TC-Net~\cite{lu2020self}. FAb-Net and TCAE followed the settings in TC-Net. Those models have all been pre-trained on the combined VoxCeleb dataset~\cite{nagrani2017voxceleb}, in contrast to the static and in-the-wild EmotioNet dataset we employ in this paper. These two datasets consist of videos of interviews containing around 7,000 subjects. TC-Net$_{k=1}$ is the result of TC-Net when the time interval is 1.
As shown in Table~\ref{DISFA}, the F1 score of our method outperforms all self-supervised methods. Even though the DISFA dataset has significant problems with data imbalance, the majority of our AU results are encouraging. The performance demonstrates the benefit of the relational learning module, which offers comprehensive AU correlation knowledge. Evaluations on BP4D are shown in Table~\ref{BP4D}. Without any temporal information, we still got identical results compared with recent self-supervised methods. In particular, AU2 (outer brow raiser), which is difficult to capture, has achieved remarkable results, suggesting that the local feature learning module can extract subtle AU changes effectively. As we can see from the results on BP4D, our method performs slightly worse than TCAE and TC-Net. This is because both of these methods were trained on video frames, and in addition, TC-Net results were obtained using an ensemble model that combined several independently learned encoders.

\begin{table*}[h!t]
\centering
\scriptsize
\caption{F1 score for self-supervised and supervised methods on BP4D~\cite{zhang2013high} of multiple facial action units (AUs).}
\label{BP4D}
\begin{threeparttable}
\begin{tabular}{@{}c|ccc|ccccc@{}}  
\toprule
\multirow{2}{*}{AU}&\multicolumn{3}{c|}{Supervised}&\multicolumn{5}{c}{Self-Supervised}\\ \cmidrule(l){2-4} \cmidrule(l){5-9} 
                    & AlexNet~\cite{chu2017learning}$\ast$   & DRML~\cite{zhao2016deep}$\ast$   & JAA-Net~\cite{shao2021jaa}$\ast$   & FAb-Net~\cite{wiles2018self} & TCAE~\cite{li2019self} & TC-Net~\cite{lu2020self}$\ast_{k=1}$ & TC-Net~\cite{lu2020self}$\ast$ & RRL(ours) \\ \midrule
1                   & 40.3     & 36.4   & 47.2      & 33.4    & 33.5 & 35.2         & \underline{42.3}   & 42.0 \\
2                   & 39.0      & 41.8   & 44.0      & 24.8    & 32.2 & 25.5         & 24.3   & \underline{35.7} \\
4                   & 41.7      & 43.0   & 54.9      & 41.0    & 43.8 & 30.2         & \underline{44.1}   & 34.0 \\
6                   & 62.8      & 55.0   & 77.5      & 73.5    & \underline{73.7} & 71.3         & 71.8   & 67.4 \\
7                   & 54.2      & 67.0   & 74.6      & 66.2    & 67.7 & 69.6         & \underline{70.5}   & 67.8 \\
10                  & 75.1      & 66.3   & 84.0      & 78.8    & 80.1 & \underline{81.3}         & 77.6   & 79.1 \\
12                  & 78.1      & 65.8   & 86.9      & \underline{84.7}    & 81.5 & 83.3         & 83.3   & 80.6 \\
14                  & 44.7      & 54.1   & 61.9      & 57.9    & 57.4 & 59.1         & 61.2   & \underline{63.9} \\
15                  & 32.9      & 33.2   & 43.6      & 21.2    & 26.5 & 30.3         & \underline{31.6}   & 28.6 \\
17                  & 47.3      & 48.0   & 60.3      & 55.7    & 54.5 & \underline{56.1}         & 51.6   & 48.6 \\
23                  & 27.3      & 31.7   & 42.7      & 26.8    & 23.2 & 27.0         & \underline{29.8}   & 26.5 \\
24                  & 40.1      & 30.0   & 41.9      & 37.9    & 31.8 & 33.4         & \underline{38.6}   & 32.4 \\
\midrule
Avg.                & 48.6      & 48.3   & 60.0      & 50.2    & 50.5 & 50.2         & \underline{52.0}   & $50.2$ \\ \bottomrule
\end{tabular}
\begin{tablenotes}
\item $\ast$ means that the values are reported in the original papers.
\end{tablenotes} 
\end{threeparttable}
\end{table*}

\textbf{Comparison with supervised methods}
We also compare our method with state-of-the-art supervised methods, including AlexNet~\cite{chu2017learning}, DRML~\cite{zhao2016deep}, and JAA-Net~\cite{shao2021jaa}. On top of the encoder, we simply train a straightforward linear classifier using frozen weights. Experimental results shown in Table~\ref{DISFA} and Table~\ref{BP4D} demonstrate that our RRL is comparable to fully supervised models. It outperforms AlexNet and DRML in BP4D and outperforms DRML in DISFA datasets. It lags behind JAA-Net, which adopts facial landmarks to jointly AU detection and face alignment. Besides, all supervised results are directly collected from the original work.

\subsection{Ablation Studies}
In this section, we conduct ablation experiments on BP4D dataset to explore the effect of each component in our proposed model and find the best structural configurations of the framework.
\vspace{-2mm}
\begin{table}[h!t]
\centering
\scriptsize
\tabcolsep=0.4cm
\renewcommand\arraystretch{1.0}
\caption{Evaluation of different components.}
\begin{tabular}{lllll}
\hline
Global & \checkmark & \checkmark & \checkmark & \checkmark \\
Local & \XSolidBrush & \checkmark & \XSolidBrush & \checkmark \\
Correlation & \XSolidBrush & \XSolidBrush & \checkmark & \checkmark \\ \hline
F1-score & 48.0 & 48.9 & 49.5 & 50.2 \\ \hline
\end{tabular}
\label{sample-table1}
\vspace{-1mm}
\end{table}

\textbf{The Effectiveness of different components:}
As shown in Table \ref{sample-table1}, Global is a baseline method that shows the results of comparative learning of only two augmented images. The F1-score rises by $0.9\%$ after the implementation of local learning, indicating the significance of local features in AU detection. By enhancing the AU relation learning via OT, the model improves the baseline by $1.5\%$. The performance demonstrates the necessity of taking into account the unique characteristics of AUs while developing self-supervised tasks. Moreover, the average F1-score is further improved by $2.2\%$ compared to the baseline when we combine those two components. Overall, both local learning and relation learning components are essential for improving the performance of the RRL network.

\begin{table}[h!t]
\centering
\scriptsize
\tabcolsep=0.4cm
\renewcommand\arraystretch{1.0}
\caption{Evaluation of the Swin Transformer and augmentation.}
\begin{tabular}{lllll}
\hline
VGG-16 & \checkmark & \XSolidBrush & \XSolidBrush \\
Swin-T & \XSolidBrush & \checkmark & \checkmark \\
Augment & \XSolidBrush & \XSolidBrush & \checkmark \\ \hline
F1-score & 48.8 & 49.1 & 50.2 \\ \hline
\end{tabular}
\label{sample-table2}
\end{table} 

\textbf{The Effectiveness of Swin Transformer and Augmentation:} 
To investigate the effect of the model backbone on the performance of AU detection, we replace the Swin Transformer backbone with VGG16 and report the results in Table \ref{sample-table2}. We can find that the use of the Swin Transformer significantly improves the detection performance because the hierarchical architecture in Swin-T extracts rich face representations. When the image augmentation is discarded, we observe that the results decline by $1.1\%$, indicating that data augmentation is crucial to self-supervised learning in natural images.


\section{CONCLUSION}
\label{sec:copyright}
In this paper, we have proposed a novel self-supervised learning framework for AU detection, in which the localization and correlation of AUs are fully considered. The framework is trained on unlabeled databases. The model has better generalization on AU detection. Extensive experimental results demonstrate the superiority of our method. In the future, we will adaptively capture the correlated regions of each AU to further extend our framework for AU intensity estimation.


\vfill\pagebreak
\label{sec:refs}
\bibliographystyle{IEEEbib}
\bibliography{strings,refs}

\end{document}